# Morphological Operation Residual Blocks: Enhancing 3D Morphological Feature Representation in Convolutional Neural Networks for Semantic Segmentation of Medical Images

Chentian Li[1, 2 (✉)], Chi Ma[1], William W. Lu[1(✉)]

[1] Department of Orthopedics and Traumatology, The University of Hong Kong, Hong Kong SAR, China
`lichent16@connect.hku.hk, wwlu@hku.hk`

[2] Department of Orthopedics and Traumatology, Zhujiang Hospital, Southern Medical University,
Guangzhou 510282, China

**Abstract.** The shapes and morphology of the organs and tissues are important prior knowledge in medical imaging recognition and segmentation. The morphological operation is a well-known method for morphological feature extraction. As the morphological operation is performed well in hand-crafted image segmentation techniques, it is also promising to design an approach to approximate morphological operation in the convolutional networks. However, using the traditional convolutional neural network as a black-box is usually hard to specify the morphological operation action. Here, we introduced a 3D morphological operation residual block to extract morphological features in end-to-end deep learning models for semantic segmentation. This study proposed a novel network block architecture that embedded the morphological operation as an infinitely strong prior in the convolutional neural network. Several 3D deep learning models with the proposed morphological operation block were built and compared in different medical imaging segmentation tasks. Experimental results showed the proposed network achieved a relatively higher performance in the segmentation tasks comparing with the conventional approach. In conclusion, the novel network block could be easily embedded in traditional networks and efficiently reinforce the deep learning models for medical imaging segmentation.

## 1 Introduction

The medical imaging segmentation of CT and MRI is an important procedure for many medical imaging analyses, including quantification of disease manifestation and 3D reconstruction, etc. In the manual segmentation methods, we noticed that the shapes and morphology of the organs and tissues are important prior knowledge in medical imaging recognition and segmentation. Using this prior knowledge of anatomic shapes may be crucial in automatic medical imaging segmentation approaches. In the imaging process field, the morphological operation is a well-known method for morphological feature extraction. As the morphological operation is performed well in hand-crafted image segmentation techniques, we assumed it is also promising to design an approach to approximate morphological operation in the convolutional networks. However, using the traditional convolutional neural network as a black-box is usually hard to specify the morphological operation action. Here, we introduced and tested several 3D morphological operation residual block designs to extract morphological features in end-to-end deep learning models for semantic segmentation.

## 2 Related Work

### 2.1 Notions of Gray−level Morphological Operation and Its Convolutional Neural Network Implementations

The morphological operation is a class of imaging processing operation based on the mathematical morphology of an image [1]. It is a special nonlinear filter for morphological feature representation and applied widely in image processing areas such as denoising, pattern recognition, etc.

The basic form of the traditional gray-level morphological operation requires two key components, a structure element, and a mathematical operation. The structure element is a basic morphology template that activates the morphology matched in the image and selected for the operation. And the mathematical operation is a set of algebraic operations that finds and returns the maximum or minimum value in the activated image region.

The morphological operation could be applied in general systems that share the properties of translation invariant, increasing (preserve a signal ordering) and semi-continuous (insensitive to fine-grained features) [2], which is suitable for many imaging processing tasks. Recently many studies were trying to use the morphological operation in deep-learning based imaging processing [3-5].

However, in the traditional form of mathematical operation mentioned above, the structure element component and operation component are separated parameters that need to be pre-defined before model training, it is not easy to be implemented in the neural networks since the parameters are hard to be designed to be automatically learned in end-to-end approach. In order to achieve such an end-to-end learning framework, Masci J et al. [6] proposed using counter-harmonic mean to approximate morphological operation instead of the traditional approach.

The Mellouli D et al. [7] further proposed using the CHM based morphological operation convolutional layer and validated the method in digits recognition task and achieved higher performance than the conventional neural network models.

Above all, the previous studies are in the concept validation stage and most of the works are tested on the 2D image datasets. More application in 3D medical imaging field is needed.

## 2.2 Fully Convolutional Neural Networks for Semantic Segmentation of Medical Images

For the semantic segmentation in medical images, it could be treated as a voxel-wise classification problem. The task is relatively complicated especially when a multiple class segmentation is required. The fully convolutional neural networks (FCNN) are proposed [8] to take advantage of extracting multilevel features and retaining the spatial information at the same time.

More recently, a modified FCNN architecture, U-net [9], is a commonly used approach for semantic segmentation in medical images. The U-net is firstly proposed by Ronneberger et al. and applied for the microscopic biomedical segmentation tasks. And Çiçek et al. [10] expanded the U-net into a 3D form to proceed with the 3D kidney image segmentation.

Following the 3D U-net architecture, many studies and researches modified the 3D U-Net to improve its performance in medical imaging segmentation. In a recent state-of-the-art architecture, Kayalibay B et al. [11] used a combination of res-net and deep-supervision techniques in the 3D U-net in image segmentation. Then Isensee et al. [12] used a similar architecture and proved to be one of the leading models in the BRATS brain tumor segmentation challenge.

## 3 Approaches

### 3.1 Design of Morphological Operation Residual Blocks

In this study, two morphological operation approaches were focused and compared for the implementation of morphological operation layer in the neural network. The first approach is based on the conventional definition of morphological operation with a non-learnable isotropic structure element (non-Learnable Morphological Operation Layer), the other approach is the CHM function approximated approach with a learnable structure element (CHM Morphological Operation Layer) following the work of Mellouli et al. [7].

For the non-learnable morphological operation layer, the structure element is set as a fixed cubic structure element of which shape equals to the convolution window shape. Then, the morphological filters could be defined as:

$$M_e(w, I) = \min \{(I(w, x) - w(x)\} \tag{1}$$

$$M_d(w, I) = \max \{(I(w, x) + w(x)\} \tag{2}$$

$$M_o(w, I) = M_d(w, M_e(w, I)) \tag{3}$$

$$M_c(w, I) = M_e(w, M_d(w, I)) \tag{4}$$

where $M_e(w, I)$ and $M_d(w, I)$ is the erosion and dilation of the image $I$ with a structure element $w$, respectively. $I(w, x)$ is a domain of image $I$ where the center of the domain is on the coordinate $x$, and the domain shape is $w$. The $w(x)$ is the subset of val-ues in a domain of *w* that outside the domain of *I(w, x)*. And *min* and *max* return the minimum and maximum value of each subset corresponding to the coordinate *x* in the image *I* respectively. Then the opening and closing are defined as $M_o$ *(w, I)* and $M_c$ *(w, I)*, respectively.

For the counter harmonic mean (CHM) based morphological operation layer, the original approach set a hyperparameter *P* to control the operation type [7]. As the previous study showed that setting *P*=1 for a dilation filter and *P*=-1 for an erosion filter is suitable for most of the imaging processing tasks, in this study we defined CHM morphological. The CHM morphological filters were defined as:

$$M_e\,(I,w)(x) = \frac{(I^0 * w)(x)}{(I^{-1} * w)(x)} \tag{5}$$

$$M_d\,(I,w)(x) = \frac{(I^2 * w)(x)}{(I * w)(x)} \tag{6}$$

where $M_e$ *(I, w) (x)* and $M_d$ *(I, w) (x)* is the erosion and dilation of the image *I* with a structure element window *w*, respectively. *I* is a domain of image *I* where the center of the subset is on the coordinate *x*. The $I^2$, $I^1$, $I^0$ and $I^{-1}$, represents a voxel-wise power operation of the image *I*. And the * denotes a convolution operation. Then the opening and closing are defined as $M_o$ *(w, I)* and $M_c$ *(w, I)*, in equation (3) and (4) respectively.

In order to equally embed the morphological operation filters into the neural network without increasing the computation complexity, the morphological operation filters are integrated into parallel pathways which are inspired by the inception module architecture [13]. Besides, we noticed that the morphological operations are usually highly non-linear filters, which may cause a high risk of gradient vanishing into the network [14]. To minimize the vanishing gradient problem when training, an identity mapping and pre-activation method are used for the architecture design based on the residual block [15]. (see Fig.1.).

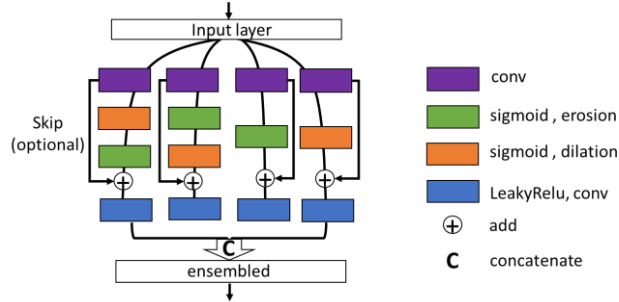

**Fig. 1.** The design of the morphological operation residual block. The morphological operation convolution layers are parallelly set in each pathway, the identity mapping and "pre-activation" are used based on the residual block. In the implementation, the erosion and dilation function could be either a choice of (1) (2) or (5) (6). And the skip connection is also an optional architecture.

In the detail of the implementation of morphological operation residual block, the basic architecture is based on a two layers CNN block. The first CNN layer is used for reducing the vector size for morphological operation. And the morphological operation layers are followed to perform morphological operations. The sigmoid function is used for pre-activation to avoid the zero values which would make the computation invalid. After the operation, the morphological filtered layer will be merged with the first CNN layer using the skip connection. A LeakyReLU activation is used after the merged layer, then the second CNN layer is used for further extraction of merged features. Finally, the morphological operation residual pathways would be ensembled together by concatenation.

### 3.2 Proposed Modified U-Net Architecture

To analysis whether the morphological operation layers can improve the medial image segmentation, in this study, we embedded the morphological operation residual block in 3D U-net like architecture proposed by Isensee et al. [12]. In the [12] architecture, the author modified the original 3D U-Net with several improvements, including deep supervision [11], and a multiclass Dice loss function [16].

In addition, the [12] architecture has a context aggregation pathway that extracting the features and a localization pathway that reconstructing the segmentation image. In order to compare with the architecture with the minimum difference, we only modified the contracting pathway by replacing the original content block to a concatenation of half channels of morphological operation residual block and half channels of the content block. The other parts were kept as same as the original [12] architecture (see Fig.2).

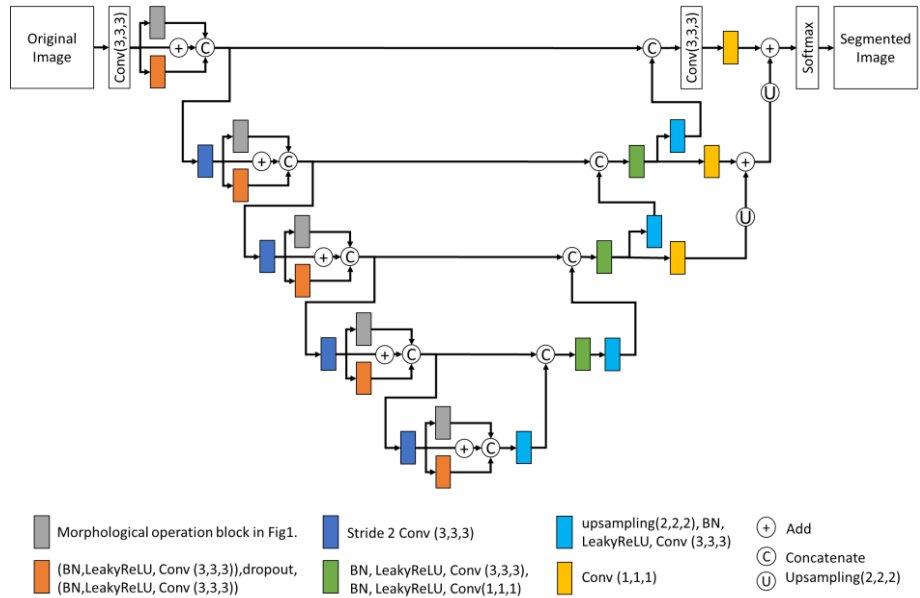

**Fig. 2.** The full CNN architecture. The implemented architecture is modified based on [12] architecture with different choice of morphological operation blocks.

## 4 Experiments and Results

### 4.1 Dataset

We compared different approaches on a brain tumor segmentation BRATS challenge old dataset from GitHub opensource data [17] [18] containing 65 LGG tumors and 102 GBM tumors cases. In the BRATS dataset, the images are manually annotated as the non-tumor, whole tumor, enhancing tumor or tumor core regions based on different modal images. The challenge goal is to segment these four regions out with multimodal images, combining T1, T2, T2-flair and T1Gd MRI sequences. All the datasets are normalized and clipped based on [12] proposed approach.

### 4.2 Implementation Details

Several modified 3D U-net architectures were built based on the architecture shown in Fig.2, which used different content and morphological operation block designs: (1) the original architecture from Isensee et al. [12]; (2) the non-learnable morphological operation block design (Fig.1.), without skip connections (non-Learnable); (3) the non-learnable morphological operation block design (Fig.1.), with skip connections(non-Learnable + skip), (4) the CHM morphological operation block design (Fig.1.), without skip connections (CHM block); (5) the CHM morphological operation block design (Fig.1), with skip connections (CHM block + skip).

The training was performed on Nvidia 1080ti GPU. All the architectures are trained with batch size 1 with early stopping strategy. The 5-fold cross-validation was performed, and Dice scores and sensitivity are computed based on the ensembled models to compare different models' performance.

### 4.3 Results

The performance of different design of morphological operation block is evaluated by the Dice score and the sensitivity.

**Table 1.** Dice Scores on the Brats Dataset.

| Model | Whole | Tumor Core | Enhancing Tumor |
|---|---|---|---|
| Isensee [12] | 0.8641 | 0.7302 | 0.6309 |
| non-Learnable | 0.8762 | 0.7572 | 0.6340 |
| non-Learnable + skip | 0.8720 | 0.7554 | 0.6430 |
| CHM Block | 0.8765 | 0.7564 | 0.6460 |
| CHM Block + skip | **0.8810** | **0.7745** | **0.6727** |

Firstly, Table 1 compared the segmentation accuracy of each architecture by Dice score. For the whole tumor segmentation, the morphological operation models were performed relatively higher than the baseline model, and the CHM with skip block reached the best performance. For the tumor core segmentation, the morphological blocks are all performed better than the baseline model, and the CHM with skip block highly increased the segmentation accuracy. Notably, for the enhancing tumor segmentation, only the CHM with skip block achieved significantly increased performance than the baseline model.

**Table 2.** Sensitivity Performance on the Brats Dataset.

| Heading level | Whole | Tumor Core | Enhancing Tumor |
|---|---|---|---|
| Isensee [12] | 0.8679 | 0.7533 | 0.7102 |
| non-Learnable | **0.8897** | **0.7983** | **0.7618** |
| non-Learnable + skip | 0.8888 | 0.7941 | 0.7628 |
| CHM Block | 0.8871 | 0.7775 | 0.7261 |
| CHM Block + skip | 0.8774 | 0.7841 | 0.7593 |

In Table 2, we compared the segmentation sensitivity of each architecture. All the four of our architectures showed higher sensitivity than the original architecture in both whole, tumor core and enhancing tumor segmentation tasks. Intriguingly, comparing the morphological operation block designs, all these four designs have similar sensitivity on the segmentation tasks, and the non-Learnable morphological operation block without skip connection is already performed well in all the three types of region segmentation.

## 5      Conclusions and Discussion

This study proposed several novel morphological operation residual block designs for medical imaging segmentation. The two types of morphological operation layer design are based on the conventional morphological operation definition and a counter harmonic mean based approximation of morphological operation. In this study, we further designed parallel pathways and residual block-like architecture for better application of the morphological operation layer.

In the experiments, we found that the counter harmonic mean based morphological operation layer blocks with skip connections achieved the best performance in the brain tumor semantic segmentation task. It shows that the design of the morphological operation layer block is effective and has a potential capacity to enhance the segmentation performance in the other 3D deep convolution neural networks. However, since the dataset and architecture tested in this study are few, the further investigation in other 3D segmentation neural networks and datasets need to be performed.

## References

1. Davidson J L, Ritter G X. Theory of morphological neural networks[C]//Digital Optical Computing II. International Society for Optics and Photonics, 1215: 378-389 (1990)

2. Maragos P. A representation theory for morphological image and signal processing[J]. IEEE Transactions on Pattern Analysis and Machine Intelligence, 11(6): 586-599. (1989)
3. Cheimariotis G A, Riga M, Toutouzas K, et al. Automatic Characterization of Plaques and Tissue in IVOCT Images Using a Multi-step Convolutional Neural Network Framework[C]//World Congress on Medical Physics and Biomedical Engineering 2018. Springer, Singapore, 261-265(2019)
4. Ball J E, Wei P. Deep Learning Hyperspectral Image Classification using Multiple Class-Based Denoising Autoencoders, Mixed Pixel Training Augmentation, and Morphological Operations[C]//IGARSS 2018-2018 IEEE International Geoscience and Remote Sensing Symposium. IEEE, 6903-6906(2018)
5. Dimitrievski M, Veelaert P, Philips W. Learning morphological operators for depth completion[C]//International Conference on Advanced Concepts for Intelligent Vision Systems. Springer, Cham,450-461(2018)
6. Masci J, Angulo J, Schmidhuber J. A learning framework for morphological operators using counter–harmonic mean[C]//International Symposium on Mathematical Morphology and Its Applications to Signal and Image Processing. Springer, Berlin, Heidelberg, 329-340(2013)
7. Mellouli D, Hamdani T M, Sanchez-Medina J J, et al. Morphological Convolutional Neural Network Architecture for Digit Recognition[J]. IEEE transactions on neural networks and learning systems (2019)
8. Long J, Shelhamer E, Darrell T. Fully convolutional networks for semantic segmentation[C]//Proceedings of the IEEE conference on computer vision and pattern recognition. 3431-3440(2015)
9. Ronneberger O, Fischer P, Brox T. U-net: Convolutional networks for biomedical image segmentation[C]//International Conference on Medical image computing and computer-assisted intervention. Springer, Cham, 234-241(2015)
10. Çiçek Ö, Abdulkadir A, Lienkamp S S, et al. 3D U-Net: learning dense volumetric segmentation from sparse annotation[C]//International conference on medical image computing and computer-assisted intervention. Springer, Cham, 424-432(2016)
11. Kayalibay B, Jensen G, van der Smagt P. CNN-based segmentation of medical imaging data[J]. arXiv preprint arXiv:1701.03056 (2017)
12. Isensee F, Kickingereder P, Wick W, et al. Brain tumor segmentation and radiomics survival prediction: contribution to the BRATS 2017 challenge[C]//International MICCAI Brainlesion Workshop. Springer, Cham, 287-297 (2017)
13. Szegedy C, Liu W, Jia Y, et al. Going deeper with convolutions[C]//Proceedings of the IEEE conference on computer vision and pattern recognition. 1-9 (2015)
14. Goodfellow I, Bengio Y, Courville A. Deep learning[M]. MIT press, (2016)
15. He K, Zhang X, Ren S, et al. Identity mappings in deep residual networks[C]//European conference on computer vision. Springer, Cham, 630-645 (2016)
16. Milletari F, Navab N, Ahmadi S A. V-net: Fully convolutional neural networks for volumetric medical image segmentation[C]//2016 Fourth International Conference on 3D Vision (3DV). IEEE, 565-571(2016)
17. Older version BRATS data , https://app.box.com/shared/ static/l5zoa0bjp1pigpgcgakup83pzadm6wxs.zip, last accessed 2019/01/29.
18. Older version BRATS data , https://app.box.com/shared/ static/x75fzof83mmomea2yy9kshzj3tr9zni3.zip,  last accessed 2019/01/29.

9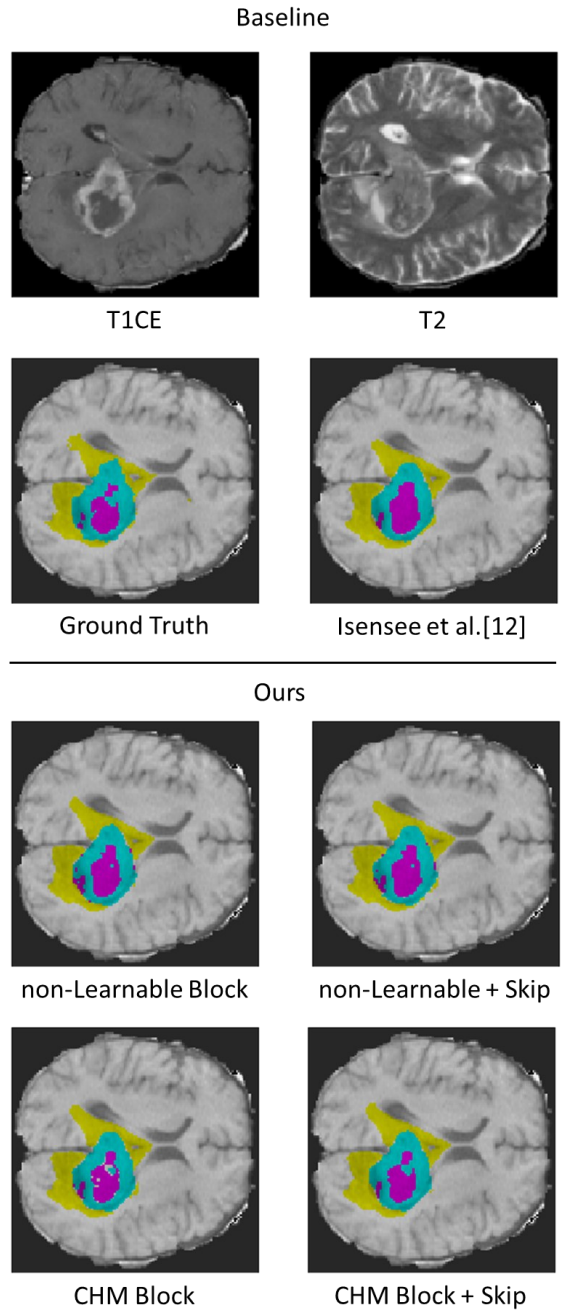

**Supplemental Fig. 2.** Representative Examples of the Brain Tumor Segmentation with different CNN model. (Yellow Label: Tumor Edema, Cyan Label: Enhancing Tumor, Purple Label: non-Enhancing Tumor Core).